\documentclass[journal,twoside,web]{ieeecolor}
\usepackage{xcolor} 
\usepackage{generic}
\usepackage{cite}
\usepackage{amsmath,amssymb,amsfonts}
\usepackage{graphicx}
\usepackage{algorithm}
\usepackage{hyperref}
\usepackage{algpseudocode}
\usepackage{multirow}
\usepackage{placeins}
\usepackage{svg}
\usepackage{stfloats} 

\hypersetup{
    colorlinks=true,
    linkcolor=black,
    citecolor=black,
    filecolor=black,
    urlcolor=black,
    pdfborder={0 0 0},
}

\def\BibTeX{{\rm B\kern-.05em{\sc i\kern-.025em b}\kern-.08em
    T\kern-.1667em\lower.7ex\hbox{E}\kern-.125emX}}
    
\begin{document}

\title{A Generative Framework for Predictive Modeling of Multiple Chronic Conditions Using Graph Variational Autoencoder and Bandit-Optimized Graph Neural Network}

\author{Julian~Carvajal~Rico, Adel~Alaeddini, Syed~Hasib~Akhter~Faruqui, Susan~P~Fisher-Hoch, and Joseph~B~Mccormick.
\thanks{© 2025 IEEE. Personal use of this material is permitted. Permission
from IEEE must be obtained for all other uses, in any current or future
media, including reprinting/republishing this material for advertising or
promotional purposes, creating new collective works, for resale or
redistribution to servers or lists, or reuse of any copyrighted
component of this work in other works.
}
\thanks{Julian~Carvajal~Rico is with the Department of Mechanical Engineering, The University of Texas at San Antonio, San Antonio, 78249 TX USA (e-mail:
julian.carvajal@utsa.edu).}
\thanks{Adel~Alaeddini is with 
the Department of Mechanical Engineering, Southern Methodist University, Dallas, 75205 TX USA (e-mail: aalaeddini@smu.edu).}
\thanks{Syed~Hasib~Akhter~Faruqui is with 
the Engineering Technology Department, Sam Houston State University, Houston, 77340 TX USA (e-mail: shf006@shsu.edu).}
\thanks{Joseph~B~Mccormick and Susan~P~Fisher-Hoch are with 
UTHealth Houston, The University of Texas Health Science Center at Houston, Houston, 77030 TX USA (e-mail: Joseph.B.McCormick@uth.tmc.edu; Susan.P.Fisher-Hoch@uth.tmc.edu).}
}

\maketitle

\begin{abstract}
Predicting the emergence of multiple chronic conditions (MCC) is crucial for early intervention and personalized healthcare, as MCC significantly impacts patient outcomes and healthcare costs. Graph neural networks (GNNs) are effective methods for modeling complex graph data, such as those found in MCC. However, a significant challenge with GNNs is their reliance on an existing graph structure, which is not readily available for MCC. 
To address this challenge, we propose a novel generative framework for GNNs that constructs a representative underlying graph structure by utilizing the distribution of the data to enhance predictive analytics for MCC. Our framework employs a graph variational autoencoder (GVAE) to capture the complex relationships in patient data. This allows for a comprehensive understanding of individual health trajectories and facilitates the creation of diverse patient stochastic similarity graphs while preserving the original feature set. These variations of patient stochastic similarity graphs, generated from the GVAE decoder, are then processed by a GNN using a novel Laplacian regularization technique to refine the graph structure over time and improves the prediction accuracy of MCC. A contextual Bandit is designed to evaluate the stochastically generated graphs and identify the best-performing graph for the GNN model iteratively until model convergence. We validate the performance of the proposed contextual Bandit algorithm against $\varepsilon$-Greedy and multi-armed Bandit algorithms on a large cohort ($n = 1,592$) of patients with MCC. These advancements highlight the potential of the proposed approach to transform predictive healthcare analytics, enabling a more personalized and proactive approach to MCC management.

\end{abstract}


\section{Introduction}
\label{sec:introduction}
\IEEEPARstart{G}{raph-based} models have become increasingly prevalent in machine learning due to their ability to model complex relational data. These models find utility in various domains, including social network analysis \cite{fan_graph_2019}, informatics \cite{wu_comprehensive_2021}, and particularly in healthcare, such as drug discovery \cite{kearnes_molecular_2016, sun_graph_2019}, molecular design \cite{dang_graph-based_2021}, and predicting disease progression \cite{zhou_graph_2021}. Although longevity has increased \cite{Beltran2015}, it has also led to an increase in chronic conditions such as diabetes, hypertension, obesity, cognitive impairment, hyperlipidemia, etc. \cite{li2020healthy, kuna2013long, shimada2019reversible, kivipelto2018lifestyle}. These chronic conditions pose serious health risks; diabetes can lead to complications like renal failure, cardiovascular disease, and vision loss \cite{tomic_burden_2022}; obesity negatively impacts mental and physical health, increasing the risk of heart disease, stroke, and several cancers \cite{avila_overview_2015, weschenfelder_physical_2018}; cognitive impairment reduces independence and raises the risk of physical injury \cite{ward_challenges_2020}; high cholesterol levels, known as hyperlipidemia, are major contributors to heart disease and stroke \cite{alloubani_relationship_2021}, and hypertension can damage the kidneys and is a leading cause of heart disease and stroke \cite{weldegiorgis_impact_2020}. These chronic conditions typically result in a lower quality of life, greater dependence on healthcare services, and a decreased ability to perform daily activities. 

The study of multiple chronic conditions (MCC) has extensively explored the impact of various risk factors on MCC networks through advanced statistical and machine learning models. Latent regression Markov clustering has identified transitions between chronic conditions using large datasets from the Department of Veteran Affairs \cite{alaeddini_mining_2017}. Temporal Bayesian networks have mapped MCC emergence and patient-level risk factors, utilizing longest-path algorithms to reveal likely comorbidity sequences \cite{faruqui_mining_2018}. Functional continuous-time Bayesian networks have modeled the influence of external variables and lifestyle behaviors on MCC development \cite{faruqui_functional_2021, faruqui_dynamic_2021}. Bayesian networks with tree-augmented naive Bayes algorithms have further identified risk factors for hepatocellular carcinoma after hepatectomy \cite{cai_analysis_2015}. These approaches illustrate the potential of advanced methodologies to elucidate the complex dynamics driving MCC.

In recent years, graph-based machine learning has evolved from traditional hand-engineered feature approaches to methods that learn node representations directly from graph structures \cite{aggarwal_node_2011, kipf_semi-supervised_2017}. This transformation is largely due to the emergence of graph-based learning, i.e., graph autoencoders (GAE) and graph variational autoencoders (GVAE) \cite{kipf_variational_2016, guo_systematic_2023}. GAE and GVAE embed features as node vectors in a low-dimensional space (latent space), allowing the storage of high-dimensional structural information to be used as features for various downstream tasks. Techniques such as random walk strategies \cite{perozzi_deepwalk_2014}, matrix factorization \cite{rohe_spectral_2011}, and GNNs have proven particularly effective in encoding and decoding schemes for node representation. Empirical evidence demonstrates the power of GVAE in tasks such as link prediction and clustering of nodes, relying on graph convolutional networks (GCN) to learn vector space representations of nodes \cite{salha_keep_2019}. These approaches have achieved competitive performance in various real-world applications, including widely used citation networks such as Cora \cite{mccallum_automating_2000}, Citeseer \cite{giles_citeseer_1998}, and Pubmed \cite{pubmed}, which serve as benchmarks for evaluating GAE and GVAE models \cite{salha_gravity-inspired_2019}.

Recent advances in the application of machine learning to healthcare have demonstrated significant potential in disease progression prediction and treatment optimization. Carvajal-Rico et al. \cite{carvajal_rico_laplacian_2024} have further advanced the application of GNNs in healthcare by incorporating Laplacian regularization. This technique enhances the learning process by ensuring similar patient nodes have similar representations, thereby improving the prediction accuracy of models dealing with MCC. Baucum et al. \cite{baucum_improving_2021} introduced transitional variational autoencoders (tVAEs) to improve reinforcement learning (RL) agents' performance by generating realistic patient condition trajectories, specifically enhancing the training of medication dosage policies through on-policy RL in intensive care settings. Kmetzsch et al. \cite{kmetzsch_disease_2022} applied Supervised VAE to predict disease progression scores using multimodal imaging and microRNA data, demonstrating improved accuracy and interpretability in patient outcome predictions. Moreover, Sun et al. \cite{sun_disease_2021} highlighted the challenges of using machine learning for disease prediction due to the reliance on abundant manually labeled EMR data, which is often insufficient for rare diseases. Their innovative model leverages external knowledge bases to augment the EMR data and uses GNNs to create highly representative node embeddings for patients, diseases, and symptoms. By aggregating information from connected neighbor nodes, their neural graph encoder can generate embeddings that capture knowledge from both data sources, allowing accurate prediction of general and rare diseases.

Building on this, our research addresses a specific gap in the use of graph-based models for healthcare applications. Traditional empirical methods fail to efficiently construct patient graphs from electronic health records (EHRs) \cite{sun_graph_2019} effectively. This is a critical step in leveraging GNNs for healthcare analytics. To overcome this limitation, we introduce a novel generative framework designed for generating and optimizing patient graphs, thus enhancing the quality of clinical data representations fed into GNN models. Our approach aims to refine patient graph representations iteratively, improving the predictive accuracy of GNNs for chronic conditions predictions. This not only advances the precision of healthcare analytics but also provides a scalable solution to the complex challenges of patient data analysis, marking a significant step forward in the application of graph-based models in healthcare.

\noindent In this work, we propose an innovative framework combining a GVAE, a Laplacian regularized graph neural network (LR-GNN), and contextual bandit (CB) algorithms to optimize graph generation from electronic health records for enhanced predictive modeling of multiple chronic conditions. Our main contributions are:

\begin{enumerate}
    \item \textbf{Generative Framework for GNNs}: We introduce a novel generative approach for constructing representative graph structures specifically designed to enhance predictive analytics for multiple chronic conditions (MCC). This framework leverages a GVAE to capture the complex relationships within patient data, facilitating a detailed understanding of individual health trajectories and generating diverse patient stochastic similarity graphs that preserve the original feature set.
    
    \item \textbf{Laplacian Regularization in GVAE}: The framework integrates the stochastically generated similarity graphs from the GVAE into an LR-GNN. This technique is applied to the generated graphs iteratively, refining the graph structure and significantly improving the prediction accuracy for MCC development.
    
    \item \textbf{Iterative Optimization with Contextual Bandit}: We design a contextual Bandit mechanism to evaluate and optimize the performance of the generated graphs iteratively. This approach assesses the predictive accuracy of each stochastically generated graph and continuously identifies the best performing graph for the GNN model until convergence is achieved, ensuring the selection of the most effective graph structures over time.
\end{enumerate}

The remainder of the paper is organized as follows: Section \ref{sec:proposed_methodology} presents the details of the proposed methodology, including the GVAE, the generation of graph variants for better model analysis, the LR-GNN used for the graph evaluation, and reinforcement learning methods for balancing exploration and exploitation of the graph variant generation. Section \ref{sec:numerical_discussion} discussed the study population, analysis of results, statistical findings, Section \ref{sec:discussion} is the discussion and limitations of the study, finally, Section \ref{sec:conclusions} provides the conclusions.

\section{Proposed Methodology}
\label{sec:proposed_methodology}
We propose a novel framework that integrates a graph variational autoencoder (GVAE), a LR-GNN, and a CB for optimal graph selection. This integration aims to refine and enhance graph representation before training the GNN model, addressing the importance of graph structure in GNN performance \cite{jin_graph_2020}. Our approach leverages GVAEs to represent graph features in a latent space, where node representations are functions of connectivity and features (detailed in Section \ref{subsec:graph_variational_autoencoder}). GVAE uses the dot product of latent representations to assess node similarity \cite{bank_autoencoders_2021, berahmand_autoencoders_2024}, removing edges with low similarity to improve the graph structure \cite{carvajal_rico_laplacian_2024}. 

The GVAE includes a sampling layer for generating various latent features, enhancing model generalization (see Section \ref{subsubsec:sampling_KL}). The reconstructed adjacency and feature matrices from the GVAE are further refined by adding Gaussian noise, creating distinct graph variants for the GNN (refer to Section \ref{subsec:graph_variants}).
The GNN uses Laplacian regularization, ensuring that similar nodes have similar representations, which improves the accuracy of MCC prediction \cite{carvajal_rico_laplacian_2024}. This process is detailed in Section \ref{subsubsec:laplacian}. The CB algorithm iteratively evaluates and selects the best-performing graphs, optimizing the graph selection process to enhance the LR-GNN model’s performance (see Section \ref{subsec:reinforcement_learning}).

\begin{figure*}[hbt!]
\centering
\includegraphics[scale=0.33]{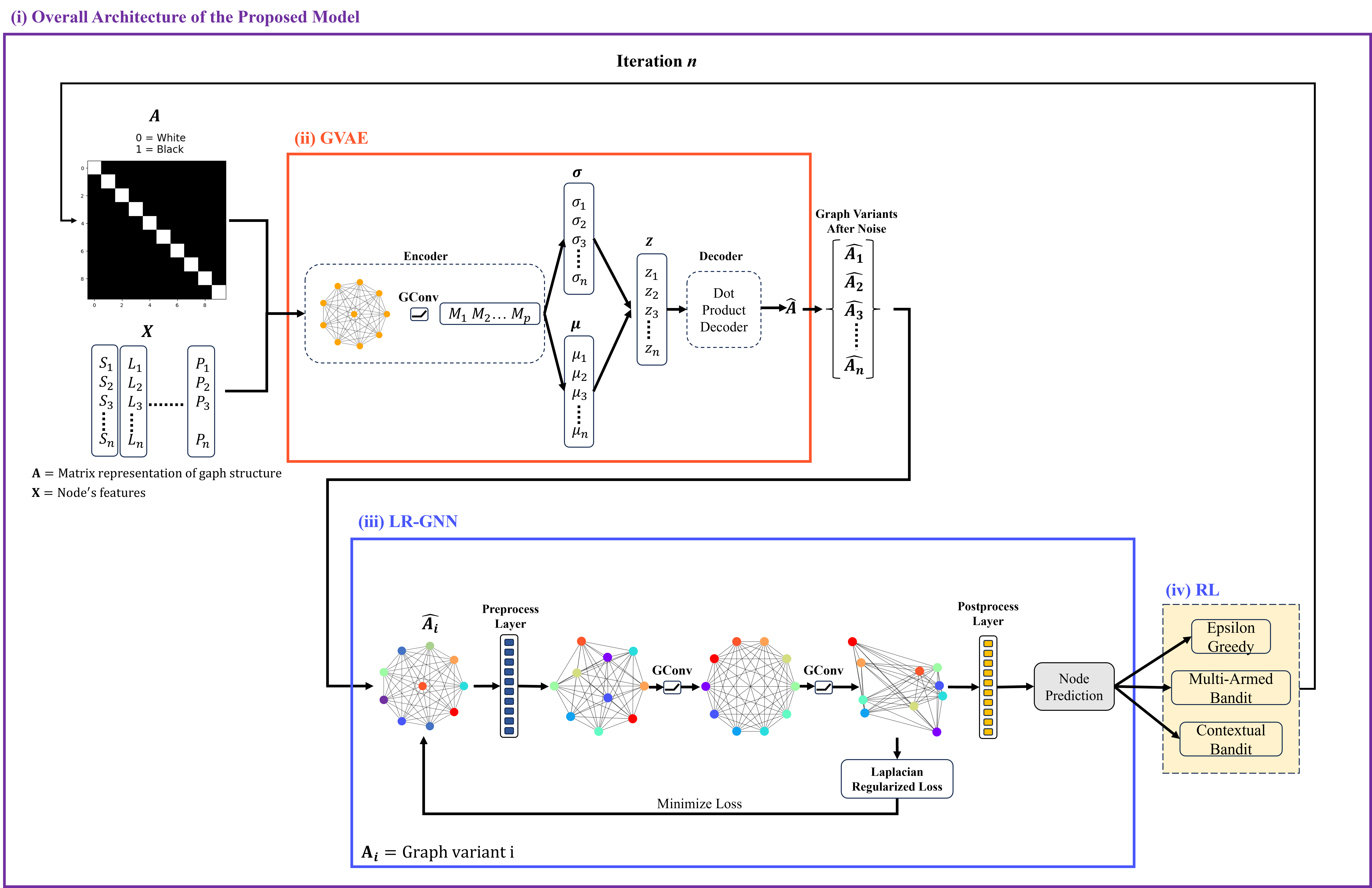}
\caption{(i) Schematic diagram of the overall proposed model. The process starts with a fully connected graph defined by its adjacency matrix $A$ and feature set $X$. The GVAE generates various graph variants, which are then used by the LR-GNN to train and evaluate MCC prediction accuracy. CB is used to identify the best performing graph, and this iterative process is repeated for comprehensive optimization across \textbf{$m$} iterations. (ii) The GVAE Architecture, where the adjacency matrix \textbf{A} and feature information \textbf{X} are used as inputs. The encoder applies graph convolutional layers to compress the graph information into a latent space representation, represented by mean $\mu$ and variance $\sigma$. Random sampling is performed in the latent space to generate latent variable \textbf{z}. The decoder then reconstructs the graph structure \textbf{$\hat{A}$}. Finally, controlled Gaussian noise is added to the reconstructed graph to generate variants for further processing by the LR-GNN. (iii) LR-GNN generates graph variants, $\hat{A}$ from the GVAE, evaluating their accuracy in predicting MCC and using reinforcement learning models ($\varepsilon$-G, MAB, CB) to select the best-performing graph for iterative refinement, and finally (iv) reinforcement learning framework.}

\vspace{-10pt}
\label{figure:GVAE_LR-GNN}
\end{figure*}

In summary, our methodology combines GVAE and LR-GNN with bandit-inspired optimization to identify and refine graph structures (see Fig.\ref{figure:GVAE_LR-GNN}(i)). This approach aims to improve the quality of graph-based learning and enhance the robustness and accuracy of the GNN model.

\subsection{Graph Variational Autoencoder}
\label{subsec:graph_variational_autoencoder}
GAEs and GVAEs have recently emerged as powerful methods for embedding nodes in a graph. These models are based on an encoding-decoding framework, where the features of a graph are represented in a latent space (encoding), then, this representation is used to reconstruct the original graph structure (decoding). This approach has been successfully applied to several challenging tasks, including link prediction, clustering of nodes, and matrix completion for inference and recommendation \cite{salha_keep_2019}. These models typically rely on GNNs to encode nodes into embeddings, specifically utilizing graph convolutional networks (GCNs) for this task.

Our proposed methodology specifically employs a GVAE to learn efficient representations of graph data. By encoding the graph into a lower-dimensional latent space, and by incorporating GVAE, we capture essential information such as features and relationships between nodes (see Fig.\ref{figure:GVAE_LR-GNN}(ii)), thereby reducing the graph's complexity \cite{salha-galvan_contributions_2022,ge_graph_2021}, retaining only the most significant necessary features for subsequent tasks such as node classification using GNNs. The process leverages the power of autoencoders to identify and extract these key features, which can include node similarity, node clustering, and other topological characteristics that are fundamental to understanding the structure of the graph and complex dynamics. In addition, sampling in the latent space allows for the generation of a graph to better represent the EHRs.

During the decoding phase, the reconstructed adjacency matrix $\hat{A}$ is obtained by performing the dot product between the node's latent space representations \cite{bank_autoencoders_2021, berahmand_autoencoders_2024}. This operation inherently assigns weights to the edges, indicating the strength of connections between nodes. These weights are important for the subsequent steps, as they reflect the importance of each edge in the graph structure.

\subsubsection{Sampling in Latent Space and Kullback-Leibler (KL) Divergence}
\label{subsubsec:sampling_KL}
The encoder of the GVAE computes the mean ($\mu$) and the log-variance ($\log(\sigma^2)$) for each node in the graph. These parameters define a Gaussian distribution from which the latent representations, $z$ (Eq. \ref{eq_z}), are sampled. This sampling mechanism introduces necessary randomness into the model, reflecting the stochastic nature of real-world graphs. The mathematical representation is as follows:

\begin{equation}
\label{eq_z}
z = \mu + \exp\left(\frac{\log(\sigma^2)}{2}\right) \cdot \epsilon, \quad \epsilon \sim \mathcal{N}(0,1)
\end{equation}


Batch stochastic sampling of the latent space to generate diverse and representative latent features that capture the underlying data distribution, improving the model's ability to generalize across different graph structures \cite{ochiai_variational_2023,xu_unsupervised_2024}. In addition to sampling, KL divergence is utilized as a regularizer in the GVAE latent space. KL divergence, often described as "a measure of how one probability distribution differs from another," ensures that the latent representations generated by the GVAE remain probabilistically meaningful. In this case, it quantifies the divergence between the learned distribution, characterized by the mean \( \mu \), and log-variance, \( \log(\sigma^2) \), of the latent representations, and a prior distribution, typically a standard Gaussian \cite{prokhorov_importance_2019}. The KL divergence is computed as follows:

\begin{equation}
\text{KL}(\mu, \log(\sigma^2)) = -\frac{1}{2} \sum_{i=1}^N \left(1 + \log(\sigma_i^2) - \mu_i^2 - \exp(\log(\sigma_i^2))\right),
\end{equation}


\noindent where $N$ is the dimension of the latent space for the summation to extend the overall dimensions of the latent space. This formulation not only encourages the model to learn meaningful and diverse representations but also ensures that these representations adhere closely to the structure of the prior distribution. By minimizing the KL divergence, the GVAE achieves a balance between fitting the data and maintaining a probabilistically sound representation in the latent space. This balance is crucial for the robustness and generalization capabilities of the model, especially when dealing with complex graph structures where the underlying data distribution can be intricate and multifaceted \cite{asperti_balancing_2020}.

\subsubsection{Edge Adjustment in Decoding}
\label{subsubsec:edge_adjustment}
During the decoding phase, the reconstructed adjacency matrix $\hat{A}$ derived from the latent space is refined by adjusting the edges based on their importance, which is indicated by their weights. Edges with lower weights are considered less significant and can be pruned to enhance the model's focus on stronger, more relevant connections:

\begin{equation}
\hat{A}_{ij} = \text{sigmoid}(z_i^T z_j),
\end{equation}

\noindent where $z_i$ and $z_j$ are the latent representations of nodes $i$ and $j$. Further, to introduce variability and test the robustness of the model, Gaussian noise is added to the reconstructed adjacency matrix:

\begin{equation}
\hat{A}_{ij}' = \hat{A}_{ij} + \mathcal{N}(0, \sigma^2),
\end{equation}

\noindent where, $\mathcal{N}(0, \sigma^2)$ represents Gaussian noise. This approach introduces controlled variability to simulate real-world uncertainties, enabling the evaluation of the model's robustness and adaptability to structural changes in graph data. The goal is to ensure that the model maintains both accuracy and stability across diverse graph conditions, thereby increasing its reliability and practicality in real-world scenarios.

\subsection{Graph Variants for Enhanced Model Exploration}
\label{subsec:graph_variants}
Generation of graph variants is important for reinforcement learning algorithms such as $\varepsilon$-G, MAB, and CB within the graph analysis framework. These variants provide alternative graph structures by introducing controlled modifications, which is particularly valuable to fully leverage the capabilities of the GVAE-GNN framework. By generating and exploring multiple graph variants, we can better understand the underlying data and improve the robustness and performance of our models \cite{adjeisah_towards_2023}. 

In this refined process, a predetermined number of graph variants are produced from an original structure, with each variant undergoing alterations exclusively in its structural aspect through controlled noise, $\mathcal{N}(0, \sigma^2)$. This selective modification introduces a measure of controlled variability, creating an array of unique scenarios for detailed examination while leaving the feature set intact \cite{dou_patchmask_2022}. The level of structural noise can be adjusted, though it is initially set to a default value to maintain consistency in the variations introduced, specifically targeting non-zero elements to preserve the original density and sparsity patterns of the graph.

In the subsequent step, after noise infusion, careful steps are taken to maintain the structural integrity of each variant of the graph; this includes applying procedures to ensure symmetry and eliminate self-loops, thereby ensuring that the resulting graphs adhere to the expected characteristics of undirected and loop-free networks, in line with established graph data protocols. As a result of this approach, a collection of graph variants is generated, each differing only in structure while maintaining the original attribute of the features. These variants become the basis for further analysis, enabling a thorough investigation into how structural changes alone can impact the model's efficiency \cite{carvajal_rico_laplacian_2024}. This targeted exploration-exploitation strategy of structural variations supports the model’s ability to progressively refine and adapt, highlighting a methodical strategy that aims to discover optimal graph configurations that could potentially lead to better results for specific tasks. Thus, this process leverages a framework for exploration within graph-based environments, leveraging structural diversity to represent possible real-life scenarios in data collection scenarios while keeping feature representations consistent.

\subsection{Graph Neural Network}
\label{subsec:gnn}
GNNs are a class of artificial neural networks designed for processing data that can be represented as graphs. Unlike traditional neural networks that operate on Euclidean data, GNNs are adept at handling graph-structured data \cite{kipf_semi-supervised_2017}. This makes them particularly useful in areas where data are naturally graph-structured, such as social networks, biochemical structures, or communication networks.

The overall architecture of the Laplacian Regularized Graph Neural Network (LR-GNN), including the input of different graph variants \(\hat{A}\) generated from the GVAE and the process of selecting the best-performing graph, is described in Fig.\ref{figure:GVAE_LR-GNN}(iii). Each graph variant's accuracy is determined in terms of predicting MCC. The accuracy of each graph is stored, and once all graphs are analyzed, the data is fed into the $\varepsilon$-G, MAB, and CB models to select the best-performing graph. This selected graph is then used as input for the GVAE to generate a new set of graph variants, repeating the procedure for a fixed number of iterations to enhance the model's performance.

The preprocess block transforms initial node features into suitable representations using a feed-forward neural network (FFN), while the postprocess block further refines these representations after the graph convolution layers, preparing them for node classification (see Fig. \ref{figure:GVAE_LR-GNN}).

\subsubsection{Laplacian Regularization}
\label{subsubsec:laplacian}
\noindent The Laplacian regularization technique is integrated into our GNN model to enhance the learning process and improve node classification accuracy \cite{carvajal_rico_laplacian_2024}. This method leverages the Laplacian matrix $L$, constructed from the graph's adjacency matrix $A$ and degree matrix $D$, where $L = D - A$, to enforce smoothness in the learned node embeddings. Specifically, nodes with similar structural or feature properties are encouraged to have similar embeddings, which aligns with the inherent relationships in the graph data. The regularization term is defined as:

\begin{equation}
\mathcal{L}_{\text {reg }}=\lambda \times \operatorname{mean}\left(\left\|y_{\text {pred }}-L y_{\text {pred }}\right\|^2\right),
\end{equation}

\noindent where $\lambda$ is a regularization parameter, $y_{\text{pred}}$ represents the predicted node embeddings, and $L y_{\text{pred}}$ captures the influence of neighboring nodes. This term ensures that the embeddings respect the graph's local structure, leading to more robust predictions. The detailed development and validation of the LR-GNN model, including the design of the Laplacian regularization term, are provided by \cite{carvajal_rico_laplacian_2024}. In this study by the incorporation of the Laplacian regularization, the GNN model achieves smoother and more reliable embeddings, which contribute significantly to the overall framework's robustness and accuracy, specifically for node classification.

\subsection{Graph Variant Selection Strategies in Reinforcement Learning}
\label{subsec:reinforcement_learning}
Reinforcement learning (RL) methods are important for optimizing decision making processes, especially in environments where decisions must be continually adjusted based on new information. These methods are designed to balance exploration, the process of trying out new actions to discover their effects, and exploitation, using known actions that produce the best results \cite{wang_exploration_2019}. In our study, we leverage several RL strategies with a focus on Contextual Bandit (CB) to improve the selection of graph variants for predictive modeling. By systematically exploring the graph outcome for MCC prediction modeling, these RL techniques aim to identify the most effective graph structures, improving the overall accuracy and reliability of the GNN model. Below, we explore the specifics of the Epsilon-Greedy ($\varepsilon$-G), Multi-armed Bandit (MAB), and Contextual Bandit (CB) approaches utilized in our framework.

\subsubsection{Epsilon-Greedy}
\label{subsubsec:e_greedy}
\noindent Our study leveraged an $\varepsilon$-G strategy, crucial for balancing exploration with exploitation in the domain of reinforcement learning. The chosen epsilon value of 10\% is a deliberate trade-off, allowing for a non-trivial amount of exploration while still predominantly utilizing the most successful graph variant to date \cite{kumar_adaptive_2024}. This stochastic strategy effectively diversifies the graph selection process, mitigating the risk of premature convergence on potentially suboptimal graph representations and fostering a thorough search within the solution space.

With 90\% exploitation, our model consistently favored variants that had previously resulted in higher accuracy, which theoretically guides the selection process towards more promising candidates over time. However, it is important to note that this strategy inherently includes a degree of randomness, which can both prevent overfitting and introduce variability in the model's performance \cite{kumar_adaptive_2024}. While advantageous in the initial phase to avoid local optima, this random component can result in occasional selections that do not align with the emerging pattern of optimal graph structures, especially as the model matures.

\noindent In contrast to the methods that followed, such as the MAB and CB, the $\varepsilon$-G strategy does not adapt its exploration rate based on the progression of learning. This distinction is important, for as the learning advances, an adaptive method that reduces exploration in favor of exploitation could potentially yield a more refined model. 

\subsubsection{Multi-armed Bandit}
\label{subsubsec:mab}
In our experiment, we employed an MAB strategy to dynamically select the most promising graph variants after the initial stages of GAE training. This approach helps to effectively balance the exploration and exploitation \cite{auer_finite-time_2002} of new graph structures against the exploitation of those that produce high precision, a core principle in reinforcement learning frameworks \cite{Sutton1998}.

For each graph variant, the GNN model is trained, and its accuracy is evaluated. These accuracies represent the actual rewards in the MAB context, and the definition for the expected reward is the following:

\begin{equation}
R_i = \frac{S_i}{\max(1, N_i)} + \alpha \sqrt{\frac{\log(N + 1)}{\max(1, N_i)}},
\end{equation}

\noindent where, $R_i$ is the estimated reward for variant $i$, $S_i$ is the number of successes (times the variant was the best choice) for variant $i$, $N_i$ is the number of trials (times the variant has been tested) for variant $i$, $N$ is the total number of iterations completed so far $\alpha$ is a parameter controlling the balance between the exploitation of known variants and exploration of new variants, finally, using the $max$ function ensures that the value in the denominator is never zero to prevent a division by zero.

In our MAB setup, the performance of each graph variant is continuously monitored and updated. As the GNN model trains each variant, the variant's success counts and trial counts are dynamically adjusted based on its performance. Specifically, a variant's success count is incremented whenever its achieved accuracy surpasses the best accuracy recorded up to that point. This method not only tracks the individual performance of the variant over time, but also helps calculate its estimated reward. The estimated reward for each variant is computed by combining its success rate with an exploration bonus. This bonus, derived from the alpha parameter and the number of trials, is crucial for improving the exploration of less frequently tested variants, thereby ensuring a balance between exploration and exploitation. Subsequently, the variant with the highest estimated reward is selected for the next training round. This selection process involves updating the trial count of the chosen variant and, if its accuracy exceeds the previously established best accuracy, updating the best accuracy metric as well. This systematic approach ensures that our model adaptively improves, continuously integrating new information to optimize performance.

\subsubsection{Contextual Bandit}
\label{subsubsec:cb}
\noindent In the progression of our experiment, we adopted a CB model that uses the predictive efficacy of Lasso regression to dynamically select the most promising graph variants subsequent to the initial training stages. The CB approach augments our model's capability to balance exploration and exploitation by integrating contextual information, thereby refining the decision-making processes within our graph data analytics framework. At the core of the Contextual Bandit framework lies Lasso Regression, chosen for its predictive accuracy and feature selection capabilities, which are instrumental in handling high-dimensional data spaces and the optimization criterion is defined as:

\begin{equation}
\min_{\beta} \left\{ \frac{1}{2N} \sum_{i=1}^N (y_i - x_i^T \beta)^2 + \lambda \|\beta\|_1 \right\}
\end{equation}

\noindent here, the model optimizes the coefficients $\beta$ by balancing the trade-off between the fitting of the training data and maintaining a sparse solution, facilitated by the regularization term $\lambda |\beta|_1$. This regularization term penalizes the absolute sum of the coefficients, encouraging a model with fewer, yet more significant, predictors. The operational flow of the CB model begins with the extraction of feature vectors that succinctly capture the salient attributes of each graph variant. Inputting these vectors into the Lasso Regression model, we project the expected rewards, which are decisive in choosing the graph variants that are most likely to enhance accuracy.

\noindent These anticipated rewards, important for the selection of graph variants, are computed considering the action taken in the given context. The model defines the expected value of the reward for an action $a_t$ within the context $c_t$ using the policy function $\pi$, as:

\begin{equation}
E[r_t(a_t, c_t) | c_t, a_t = \pi(c_t)]
\end{equation}


\noindent here, $E$ denotes the expected value and $r_t(a_t, c_t)$ represents the reward received at time $t$ for choosing action $a_t$ in context $c_t$. The policy $\pi(c_t)$ is the strategy that selects the action based on the context $c_t$, which is crucial to determine the most beneficial graph variant to employ at each iteration.

\begin{figure}[ht]
\centering
\includegraphics[scale=0.48]{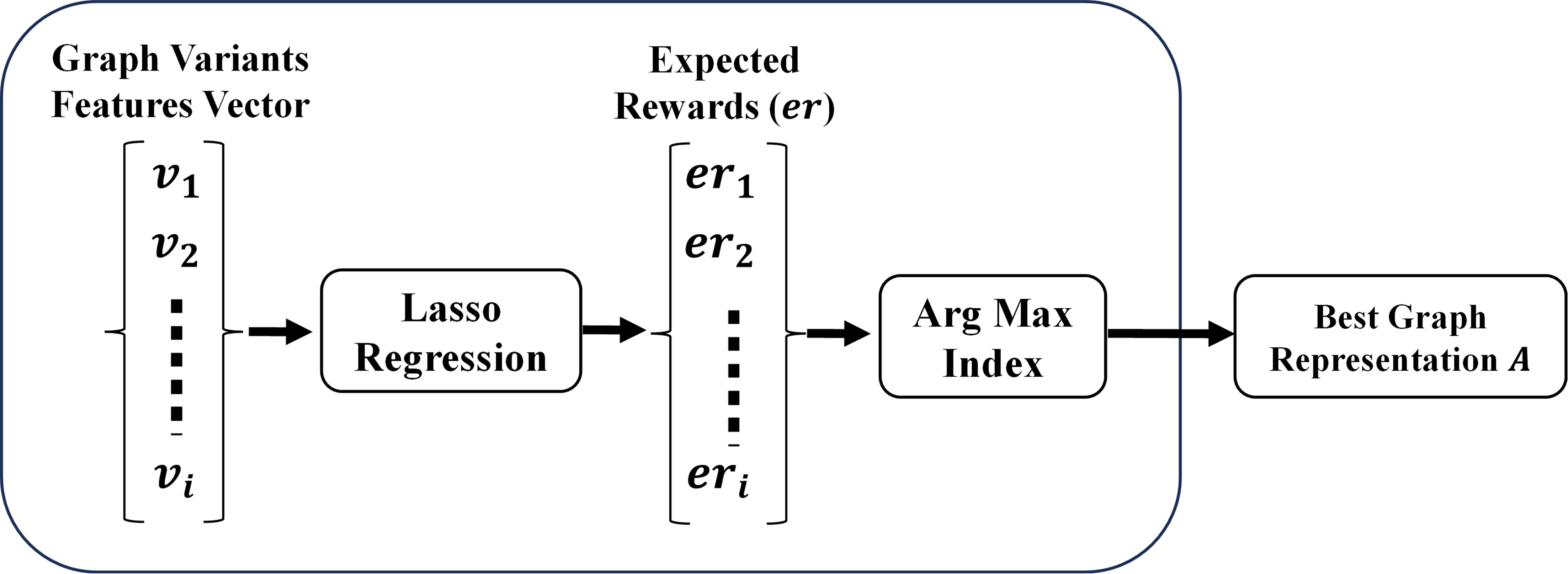}
\caption{Schematic Representation of Contextual Bandit (CB) Implementation. This figure illustrates the selection process of the best graph representation using CB. The expected reward is computed based on the graph context (features) using Lasso regression. The best graph and corresponding accuracy are then selected based on the index of the highest expected reward.}
\vspace{-10pt}
\label{figure:CB}
\end{figure}

\noindent With the selection of a variant (see Fig. \ref{figure:CB}), the CB model assimilates the results to iteratively refine its predictive capability. This process of continual learning and adaptation is indispensable for the CB approach, as it ensures that the model becomes progressively more adept at making accurate selections. Detailed documentation of the performance of each variant informs the process, fostering transparency and ongoing optimization. Incorporating this context-driven technique has marked a significant evolution in our methodological approach. It stands as a testament to our commitment to integrating sophisticated strategies that harness the power of contextual data, setting a new standard for achieving precision in graph-structured data analytics.

In summary, the framework, as illustrated in Fig. \ref{figure:GVAE_LR-GNN}, begins with a fully connected graph characterized by its adjacency matrix $A$ and the set of features $X$. This graph is processed by the GVAE to generate different graph variants. Each of these variants is then used as input for the LR-GNN to train the model and evaluate the accuracy of MCC prediction for each graph representation. Once the accuracies for all graph variants are recorded, these results are fed into reinforcement learning methods—$\varepsilon$-G, MAB, and CB, to identify the best performing graph representation. This selected graph is then used as the input for the GVAE, repeating the entire process iteratively until convergence. In this study, the complete loop is repeated 15, 20, and 25 times, allowing for comprehensive exploration and refinement of graph structures to optimize the GNN model’s performance.

\section{Numerical Results}
\label{sec:numerical_discussion}
In this section, we present the numerical results and analysis of our proposed framework. The primary objective of this section is to evaluate the performance of the GVAE-GNN model using the Cameron County Hispanic Cohort (CCHC) \cite{FisherHoch2010} (Committee for the Protection of Human Subjects: IRB HSC-SPH-03007 Bd) dataset. We evaluate the model's efficacy and efficiency by conducting a series of experiments to compare the reinforcement learning methods employed: $\varepsilon$-G, MAB, and CB. Through these experiments, we aim to identify the optimal approach for predictive modeling of MCC. The CB method was the most effective approach during this process, as it accounts for the contextual information of the graph variants while maintaining a balance between exploration and exploitation. In our implementation, context refers to the feature vectors extracted from each graph variant, which encapsulates structural properties and node attributes for predictive modeling. These features include, node embeddings, and statistical measures that characterize each graph representation. By integrating these contextual features, the CB model dynamically adapts its selection strategy, prioritizing graph variants that are more likely to improve the accuracy of the model. This ability to leverage graph-specific context allows the CB model to make more informed decisions compared to non-contextual approaches, leading to superior predictive performance. The integration of GVAE, RL, and GNN ensures a cohesive flow: (i) GVAE generates graph variants, (ii) GNN trained on the variants generates evaluation metrics for the graphs, and finally, (iii) RL evaluates and selects the best graph variant and the associated robust features for the best performing model. 
\noindent This iterative optimization ensures that the GNN leverages graph structures that maximize its predictive performance while efficiently exploring the graph space. By connecting the GVAE-generated graphs to the bandit optimization process, our framework achieves a seamless integration of generative modeling, optimization, and graph-based learning.

\noindent The following subsections provide detailed descriptions of the study population, analysis, and statistical findings.

\subsection{Study population}
\label{subsec:study_population}
\begin{table*}[b!]
\centering
\caption{Statistics of the population from CCHC dataset}
\includegraphics[scale=0.53]{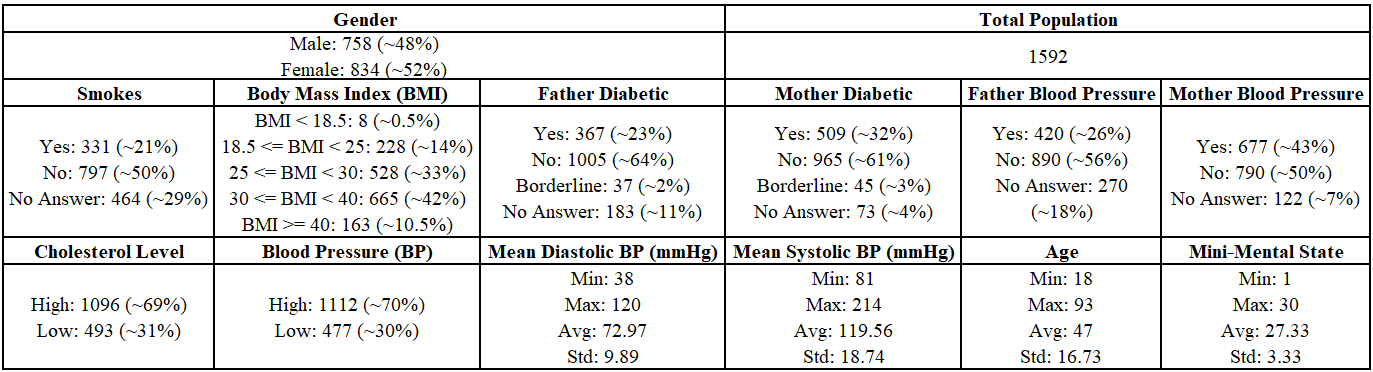}
\label{Table:Statistics}
\end{table*}
In this study, we use the CCHC dataset to validate our proposed GNN model. The CCHC is a cohort study that began in 2004, consists primarily of Mexican Americans (98\%), and currently enrolls 5,020 patients. It provides a unique opportunity to study health disparities in a community along the Texas-Mexico border. The CCHC employs a rotating recruitment strategy and uses random selection to enroll participants from this community, which has notable health disparities. The inclusion criteria for our model required participants to have valid information on risk variables such as age, body mass index (BMI), typical blood pressure, cholesterol level, smoking habits, parental history of diabetes (father/mother), parental history of hypertension (father/mother), systolic and diastolic blood pressure, mini-mental state examination (MMSE) score (a cognitive impairment indicator), and gender.

For model validation, we are focused on a subset of 1,592 patients that met the criteria of not having missing data. This selection aids in analyzing the relationship between five chronic conditions: diabetes, obesity, cognitive impairment, hyperlipidemia, and hypertension. Each patient (node) in our framework is assigned a feature vector consisting of these risk factors, enabling the model to learn associations between patients' risk factors and MCC outcomes. The LR-GNN model predicts the possible outcome for a patient by categorizing it into one of 32 possible classes, representing different combinations of the five chronic conditions. This allows the model to capture patterns of MCC rather than predicting single conditions. By leveraging graph-based learning models, we can analyze the complex interaction among risk factors and improve predictive accuracy in identifying high-risk patients.

The model inclusion criteria included valid information on risk variables for the respondents, such as age, body mass index, usual blood pressure, cholesterol level, smoking, and family medical history. The model was validated using 1,592 patients from the dataset, more specifically, those who met all the criteria and did not have missing data. Patients submitted the initial information, which was then systematically collected and confirmed by healthcare professionals, such as doctors and nurses. This method reduces the likelihood of recollection bias and other frequent mistakes related to self-reporting.

The healthcare personnel collecting data are trained to extract accurate and detailed information, ensuring the reliability of the data set. Although we recognize that no data collection approach is completely free from biases or errors, our study's mix of patient self-reports and professional scrutiny provides a solid and reliable foundation for our research. The CCHC dataset was rigorously cleaned during the development and validation of our model. 

We ensured that patient information was complete by including only records that contained complete data on all risk factors related to the five chronic conditions under investigation. To avoid modeling errors due to missing data, we carefully selected individuals with incomplete information. Table \ref{Table:Statistics} presents a summary of the CCHC data for the selected patients.

\subsection{Results Analysis}
\label{subsec:result_analysis}

\begin{figure}[tp!]
\centering
\includegraphics[scale=0.37]{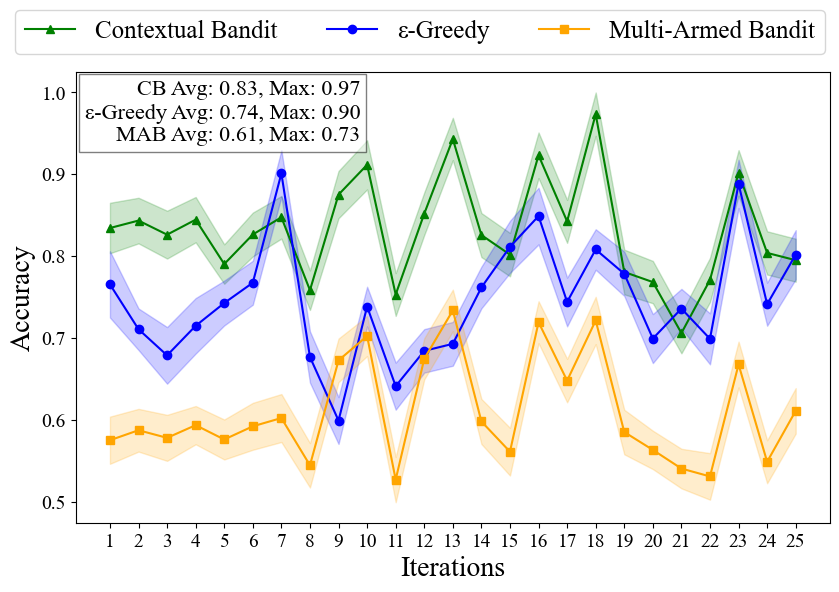}
\caption{Comparative Accuracy for Best Graph Variant Selection Methods for 25 Iterations}
\label{Figure:25_iterations}
\vspace{-10pt}
\end{figure}

To validate our model, we conducted three different tests by varying the number of iterations to train the model (i.e., 25, 20, and 15 iterations) to determine the optimal stopping point for evaluating the performance of three reinforcement learning methods: $\varepsilon$-G, MAB, and CB. These tests provided substantial insights into the performance of our integrated GVAE-GNN model for the prediction of MCC.

We initially chose 25 iterations as a starting point to provide a robust baseline for model evaluation. To explore the effects of iteration on model training, we reduced the number of iterations required for training by five, from 25 to 20. This training iteration reduction aimed to identify whether a smaller number of iterations could yield similar performance, potentially reducing computational resources and time. One additional reduction to 15 iterations was performed; however, reducing iterations further from 15 did not yield competitive results. Thus, these results are not reported in this work.

The initial set of 25 iterations demonstrated the superior performance of the CB method, as evidenced by its highest average and maximum accuracy, as shown in Fig. \ref{Figure:25_iterations}. In this figure, we also plot the 95\% confidence interval for each method based on six iterations to compare the possible worst-best scenarios for each method.

\begin{figure*}[ht]
\centering
\includegraphics[scale=0.3]{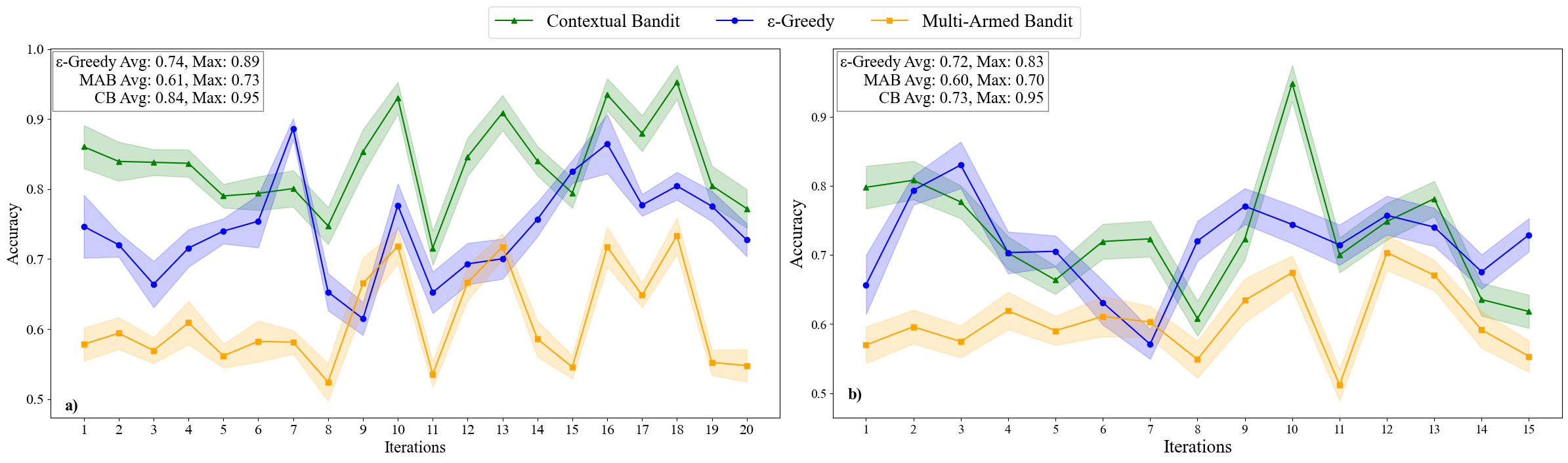}
\caption{\textbf{a)} Comparative Accuracy of 20 iteration, \textbf{b)} Comparative Accuracy of 15 iterations, for Best Graph Variant Selection Methods}
\label{Figure:20_15_iterations}
\vspace{-10pt}
\end{figure*}

Upon reducing the iterations to 20 and 15 (see Fig.\ref{Figure:20_15_iterations}), we aimed to identify an early stopping point without compromising the performance of the model. The results of the 20 iterations were remarkably close to those of the 25 iterations, suggesting that 20 iterations might suffice for reliable performance evaluation. Similarly to the 25 iterations, the 20- and 15-iteration figures also include the 95\% confidence intervals based on six iterations per method.

When comparing the metrics of 20 and 25 iterations (see Table \ref{Table:Comparison}), the performance metrics of the CB method remain consistent, with only slight variations in accuracy. The mean accuracies for the CB method are very close in both scenarios, suggesting that extending beyond 20 iterations yields marginal improvements. This stability is also reflected in the variance and inter-quartile range (IQR), which show similar values, indicating consistent performance across different iterations in Fig.\ref{Figure:box_plot}.

\subsubsection{Comparison with Classic Machine Learning Models}

\noindent To further evaluate the efficacy of our GVAE-GNN model, we compared it against three classic machine learning models: \textbf{K-Nearest Neighbors (KNN)}, \textbf{Logistic Regression (LR)}, and a \textbf{Feedforward Neural Network (FNN)}. Consistent preprocessing steps were applied across all methods to ensure fairness in the comparison. The results are summarized below in Table \ref{table:models comparisson}:

\begin{table}[h!]
\centering
\caption{Comparison between best performing set-up and classic non-graph based machine leanring models}
\begin{tabular}{|c|c|c|c|c|}
\hline
\textbf{Model/Metric} &\textbf{GVAE-GNN} & \textbf{KNN} & \textbf{LR} & \textbf{FNN} \\ \hline
Test Accuracy & \textbf{0.8316} & 0.2720 & 0.3013 & 0.2579 \\ \hline
F1-Score & \textbf{0.8238} & 0.2360 & 0.2337 & 0.2202 \\ \hline
Recall & \textbf{0.8240} & 0.2720 & 0.3013 & 0.2579 \\ \hline
Precision & \textbf{0.8288} & 0.2539 & 0.2106 & 0.2022 \\ \hline
\end{tabular}
\label{table:models comparisson}
\end{table}

In contrast, our GVAE-GNN model consistently outperformed all baseline models. Its ability to leverage both structural and feature information contributed to its superior results. By integrating graph-based representations and reinforcement learning, the GVAE-GNN demonstrated greater accuracy, robustness, and stability across varying iterations. 
As detailed in the comparative analysis, KNN, Logistic Regression, and FNN all faced significant challenges in handling the dataset's multi-class nature and high feature dimensionality. While these methods offer simplicity and computational efficiency, their inability to capture nuanced relationships in the data led to suboptimal performance. In contrast, the GVAE-GNN combined with the CB approach not only outperformed these models but also demonstrated greater flexibility and adaptability, reinforcing its suitability for MCC prediction.

\subsection{Statistical Analysis}
\label{subsec:statistical_findings}

The statistical analysis of the methods across different iterations provides deeper insights into their performance and reliability. The $\varepsilon$-G approach demonstrated potential, with an accuracy of 90.06\% for 25 iterations, 88.62\% in the seventh iteration for 20 and 83.04\% for 15 iterations. However, the inherent randomness in the $\varepsilon$-G method, which sometimes favors exploration over exploitation, does not guarantee the best outcome in every iteration. This randomness, while useful in the early stages of a broad exploration, proved to be a limitation when stability and reliability are essential.

\begin{table*}[b]
\centering
\caption{Comparison of methods based on various metrics for 15, 20, and 25 iterations}
\scriptsize
\resizebox{0.9\textwidth}{!}{
\begin{tabular}{|c|c|c|c|c|c|c|c|c|c|c|}
\hline
\multirow{2}{*}{Metric} & \multicolumn{3}{c|}{15 Iterations} & \multicolumn{3}{c|}{20 Iterations} & \multicolumn{3}{c|}{25 Iterations} \\ \cline{2-10}
 & $\varepsilon$-G & MAB & CB & $\varepsilon$-G & MAB & CB & $\varepsilon$-G & MAB & CB \\ \hline
Mean                    & 0.7164 & 0.6036 & \textbf{0.7305} & 0.7423 & 0.6117 & \textbf{0.8369} & 0.7450 & 0.6099 & \textbf{0.8316} \\ \hline
Variance                & 0.0039 & \textbf{0.0025} & 0.0071 & 0.0047 & 0.0045 & \textbf{0.0037} & 0.0049 & \textbf{0.0039} & \textbf{0.0039} \\ \hline
Q1 (25th Percentile)    & \textbf{0.6896} & 0.5722 & 0.6818 & 0.6985 & 0.5593 & \textbf{0.7942} & 0.6987 & 0.5631 & \textbf{0.7900} \\ \hline
Q3 (75th Percentile)    & 0.7509 & 0.6268 & \textbf{0.7792} & 0.7768 & 0.6656 & \textbf{0.8651} & 0.7781 & 0.6674 & \textbf{0.8515} \\ \hline
F1-Score                & 0.7358 & 0.6615 & \textbf{0.7622}          & 0.7852 & 0.6758 & \textbf{0.8116}          & 0.7901 & 0.6633 & \textbf{0.8238}          \\ \hline
Recall                  & 0.7412 & 0.6686 & \textbf{0.7636}          & 0.7870 & 0.6814 & \textbf{0.8123}          & 0.7916 & 0.6698 & \textbf{0.8240}          \\ \hline
Precision               & 0.7549 & 0.6893 & \textbf{0.7741}          & 0.7960 & 0.6978 & \textbf{0.8180}          & 0.7968 & 0.6887 & \textbf{0.8288}          \\ \hline
AUC                     & 0.9564 & 0.9403 & \textbf{0.9675}          & 0.9673 & 0.9392 & \textbf{0.9766}          & 0.9686 & 0.9364 & \textbf{0.9779}          \\ \hline
\end{tabular}%
}
\label{Table:Comparison}
\end{table*}

In contrast, the MAB strategy, although methodical, faced challenges with the dynamic nature of the graph generation landscape. It struggled to adapt to the constant variations post-Gaussian noise infusion and GVAE applications, which is apparent from the lowest median levels over the iterations (see Table \ref{Table:Comparison}). Despite this, the MAB strategy had a maximum accuracy of 73. 34\%
when running the model for 25 iterations, then after running the model for 20 iterations it reached a similar accuracy of 73.34\%, and finally a maximum accuracy of 70.35\% for 15 iterations. This suggests that even with increasing the number of iterations, the MAB did not improve over time.

However, the transition to the contextual bandit strategy marked a significant improvement in the refinement of the model. By incorporating the specific context of each graph variant, we were able to predict expected rewards more accurately and select variants with greater potential for accuracy. Notably, the CB achieved an exceptional peak accuracy of 97.30\% for 25 iterations, 95.26\% for 20 iterations, and 94.89\% for 15 iterations, indicating the superiority of the contextual approach over the others. The emphasis of this method on the relevant features and contextual details of each graph variant led to a more strategic and less random selection process.

\section{Discussion}
\label{sec:discussion}
In this study, we compared the performance of three popular reinforcement learning (RL) algorithms: $\varepsilon$-Greedy, MAB, and CB. The results, detailed in the previous section, highlight the differences in performance among these methods across a range of iterations from 15 to 25. This incremental experimentation underscores the necessity of a sufficient number of iterations to accurately assess the performance of different methods. The CB method consistently demonstrated superior results in terms of maximum accuracy, indicating its robustness and effectiveness in handling complex data structures. The slight improvements observed between 20 to 25 iterations suggest that while additional iterations can provide marginal enhancements, they may not always justify the increased computational cost.

\begin{figure}[tp!]
\centering
\includegraphics[scale=0.22]{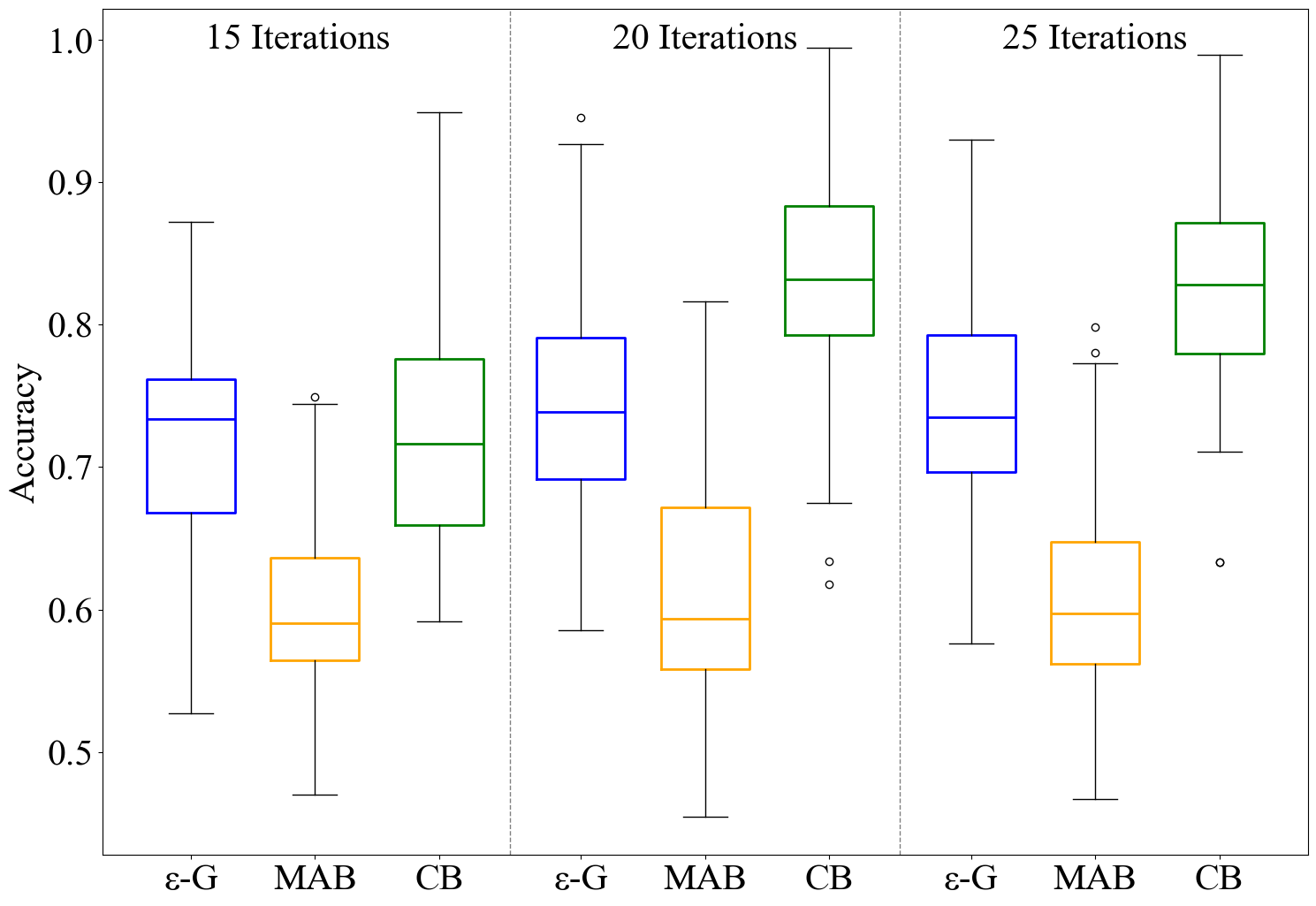}
\caption{Minimum, Maximum and Median Accuracy Over 15, 20 and 25 Iterations.}
\label{Figure:box_plot}
\vspace{-10pt}
\end{figure}

The box and whisker plots in Fig. \ref{Figure:box_plot}, illustrate the performance differences across 15, 20, and 25 iterations for the three bove-mentioned reinforcement learning methods: $\varepsilon$-G, MAB, and CB.

For 15 iterations, the $\varepsilon$-Greedy algorithm displays a median accuracy of approximately 0.72 with a moderate interquartile range (IQR), indicating consistent performance with some variability. The Multi-Armed Bandit (MAB) algorithm shows the lowest median accuracy, around 0.6, with a smaller IQR, reflecting consistently lower performance but less variability. Conversely, the Contextual Bandit (CB) algorithm achieves a median accuracy close to 0.73 with a larger IQR, suggesting better performance despite greater variability. At 20 iterations, $\varepsilon$-Greedy maintains a median accuracy of around 0.72 but exhibits a slight increase in variability. MAB continues to have the lowest median accuracy, around 0.6, with an increased IQR, indicating more variability. CB again shows the highest median accuracy, approximately 0.83, with an even larger IQR, reflecting its continued superiority despite higher variability. With 25 iterations, $\varepsilon$-Greedy shows a slight improvement in median accuracy and reduced variability, suggesting better consistency. MAB's performance remains similar to previous iterations, with low median accuracy and a small IQR. CB achieves the highest median accuracy, around 0.83, with a reduced IQR compared to previous iterations, indicating improved consistency.

Overall, the combined plots underscore the superiority of the CB method, demonstrating consistently higher median accuracies across all iterations and the highest maximum accuracies. This suggests that CB is more effective and reliable for achieving higher accuracy in predictive modeling, despite initial high variance.

These findings highlight the importance of considering not just average performance but also the variability and distribution of outcomes when evaluating a model's performance. Despite some initial variability, the CB method consistently demonstrated higher accuracy and reliability over more iterations, making it a better choice for applications requiring high accuracy and improvement over time.

To further evaluate the performance differences between $\varepsilon$-Greedy, MAB, and CB, we also conducted a p-value analysis to determine the statistical significance of the observed differences. A p-value less than 0.05 indicates a statistically significant difference between the methods, suggesting that the observed differences are unlikely to have occurred by chance. Conversely, a p-value greater than or equal to 0.05 suggests that any observed differences could be due to random variation. Table \ref{table:p_values} illustrates the p-values for each comparison and iteration count.

\begin{table}[h!]
\centering
\caption{P-values for Different Iterations}
\begin{tabular}{|c|c|c|c|}
\hline
\textbf{Comparison} & \textbf{15 Iterations} & \textbf{20 Iterations} & \textbf{25 Iterations} \\ \hline
$\varepsilon-G$ vs MAB & $1.18 \times 10^{-24}$ & $9.85 \times 10^{-29}$ & $5.28 \times 10^{-38}$ \\ \hline
$\varepsilon-G$ vs CB & $0.193$ & $5.34 \times 10^{-21}$ & $2.11 \times 10^{-23}$ \\ \hline
MAB vs CB & $2.25 \times 10^{-30}$ & $7.95 \times 10^{-77}$ & $8.30 \times 10^{-100}$ \\ \hline
\end{tabular}
\label{table:p_values}
\end{table}

For 15 iterations, the comparison between the $\varepsilon$-Greedy and MAB methods yields a p-value of $1.18 \times 10^{-24}$, indicating a statistically significant difference between the two methods. The comparison between $\varepsilon$-Greedy and CB methods results in a p-value of 0.193, suggesting no statistically significant difference between these methods, as the p-value is greater than 0.05. The comparison between the MAB and CB methods shows a p-value of $2.25 \times 10^{-30}$, indicating a statistically significant difference. For 20 iterations, the comparison between $\varepsilon$-Greedy and MAB methods produces a p-value of $9.85 \times 10^{-29}$, indicating a statistically significant difference. The comparison between $\varepsilon$-Greedy and CB methods yields a p-value of $5.34 \times 10^{-21}$, also suggesting a statistically significant difference. The comparison between the MAB and CB methods shows a p-value of $7.95 \times 10^{-77}$, indicating a statistically significant difference. For 25 iterations, the comparison between $\varepsilon$-Greedy and MAB methods results in a p-value of $5.28 \times 10^{-38}$, indicating a statistically significant difference. The comparison between $\varepsilon$-Greedy and CB methods yields a p-value of $2.11 \times 10^{-23}$, suggesting a statistically significant difference. The comparison between the MAB and CB methods shows a p-value of $8.30 \times 10^{-100}$, indicating a statistically significant difference.

These comparisons consistently show that $\varepsilon$-Greedy significantly outperforms MAB across all iteration sets. There is no significant difference between $\varepsilon$-Greedy and CB at 15 iterations. However, significant differences are observed at 20 and 25 iterations, indicating that CB outperforms $\varepsilon$-Greedy as the number of iterations increases. MAB consistently shows a statistically significant difference compared to CB across all iteration sets, with CB consistently outperforming MAB.

Looking to the future, our objective is to expand our dataset to incorporate additional variables, potentially improving the model's predictive power. Investigating more sophisticated graph learning algorithms may significantly refine our approach. The deployment of our model in real-world clinical settings remains the definitive test of its practicality and utility.

In summary, our work demonstrates that the contextual bandit approach offers a more consistent and reliable path to model optimization in healthcare analytics. By strategically using the learned graph structure and implementing advanced selection strategies, our integrated GVAE-GNN framework not only streamlines the learning process but also significantly improves predictive accuracy for complex tasks in healthcare. Furthermore, this approach has the potential to develop personalized recommendation systems, providing users with tailored suggestions based on their individual preferences. Our study highlights the importance of carefully selecting the right graph structure for the predictive GNN model. In scenarios where the context is limited or noisy, other methods such as Q-learning or SARSA may be more effective.

\vspace{-15pt}

\subsection{Model Generalizability}
\label{subsec:model_validation}
To evaluate the generalizability of our framework beyond the original study population, we performed an external validation using a more generalized synthetic dataset. This dataset was evaluated using the best-performing combination with the CB algorithm as the reinforcement learning model. Given the lack of publicly available datasets with a structure and feature set comparable to the CCHC, we created a synthetic dataset that not only preserved key variable relationships but also incorporated additional risk factors. These new features were designed to enhance model robustness by reflecting a broader range of health conditions and patient characteristics.

To ensure the synthetic dataset integrity, we conducted a distribution analysis using the \textit{distfit}\cite{Taskesen_distfit_is_a_2020} Python library on key variables such as age (Dagum distribution), systolic and diastolic blood pressure (Johnson's SU-distribution), gender (Cauchy distribution), among others. This process allowed us to generate realistic risk factor distributions while maintaining statistical consistency with the original CCHC dataset. The synthetic dataset was constructed in a way that retained essential dependencies while introducing controlled variability, ensuring a more rigorous test of the model's ability to generalize.

The models were then trained and evaluated on this synthetic dataset, with predictive performance measured using precision, recall, F1-score, and AUC-ROC, see Table \ref{table:external_validation}. These metrics provide a comprehensive assessment of the model’s capability to handle variability in healthcare data. The results of this validation confirm the model’s ability to generalize beyond the initial cohort, reinforcing the reliability of our approach for predictive modeling of MCC in diverse populations.

\begin{table}[h!]
\centering
\caption{Validating proposed model with synthetic dataset}
\begin{tabular}{|c|c|c|c|}
\hline
\textbf{GVAE-GNN} &\textbf{15 Iterations} & \textbf{20 Iterations} & \textbf{25 Iterations} \\ \hline
Avg. Test Accuracy & 0.7809 & 0.7965 & 0.8165 \\ \hline
F1-Score & 0.7813 & 0.7945 & 0.8037 \\ \hline
Recall & 0.7818 & 0.7941 & 0.8035 \\ \hline
Precision & 0.7865 & 0.8009 & 0.8023 \\ \hline
AUC & 0.9823 & 0.9839 & 0.9887 \\ \hline
\end{tabular}
\label{table:external_validation}
\end{table}

\subsection{Study Limitations and Considerations}
\label{subsec:limitations}

While our study introduces a pioneering generative GVAE-GNN for characterizing the emergence of MCC, it is important to acknowledge its limitations. Our research uses the CCHC, focusing on a specific demographic that primarily consists of Mexican Americans. This focus, while valuable for in-depth analysis, restricts the generalizability of our findings to other populations with different demographic profiles or characteristics of the healthcare system.
Additionally, the complexity of integrating generative models like GVAE into healthcare analytics involves understanding the complexity of patient data and the stochastic nature of graph generation, which might impact the model’s interpretability and real world applicability.
Incorporating Laplacian regularization and the $\varepsilon$-G, multi-armed Bandit and contextual Bandits algorithms in our GVAE-GNN model signifies an advanced methodological approach. However, these innovations also introduce computational complexity and require rigorous evaluation to ensure that the model recommendations are robust and reliable for clinical use.
Another consideration is the model's scalability and adaptability to other chronic conditions beyond the five initially studied. Although designed to accommodate additional conditions, the practicality of this extension must be carefully examined in future research, considering the potential need for substantial adjustments to the graph structure and underlying algorithms.
Lastly, integrating a generative model with a GNN presents novel challenges in model training and validation. Ensuring that the generated graphs accurately represent patient similarities and contribute meaningfully to the predictive accuracy of the model requires a balance between the model's generative and discriminative capabilities.
In conclusion, while our GVAE-GNN model offers substantial promise in advancing personalized healthcare analytics, these limitations and considerations underscore the need for careful interpretation of the model findings and deliberate planning for its integration into clinical practice.

\section{Conclusions }
\label{sec:conclusions}
This study develops a novel generative framework that combines the Graph Variational Autoencoder (GVAE) and Laplacian-regularized Graph Neural Networks (GNNs) optimized via Contextual Bandit (CB) algorithms for predictive modeling of multiple chronic conditions (MCC). By addressing the critical challenge of constructing a representative graph structure from complex patient data, our framework demonstrates significant improvements in the accuracy of MCC predictions.  GVAE effectively captures intricate relationships within patient data, facilitating the generation of diverse and representative patient similarity graphs. Integrating Laplacian regularization within the GNN further refines these graphs, ensuring the model’s robustness and enhancing its predictive capabilities. The iterative optimization process through CB algorithms significantly outperforms traditional methods such as $\varepsilon$-Greedy and multi-armed bandit approaches. 

Our extensive validation on a large cohort of patients underscores the framework’s potential to transform predictive healthcare analytics, paving the way for more personalized and proactive management of MCC. Specifically, the application of our framework to the CCHC, a minority, low-income Hispanic/Latino population (H/L; Mexican-Americans) with high rates of metabolic diseases such as diabetes (26\%) and obesity (51\%), as well as associated complications like cardiovascular and liver diseases, demonstrates its clinical relevance. By effectively modeling the complex interplay of chronic conditions in this high-risk population, our approach has the potential to significantly improve early detection and management strategies, ultimately reducing the burden of these diseases \cite{FisherHoch2015}.

These advancements highlight the critical role of innovative graph machine learning techniques in addressing complex healthcare challenges. Future research will explore the scalability of our approach in different populations and conditions, with the aim of further evaluating and enhancing its applicability in diverse clinical settings. By continually refining graph structures and incorporating additional contextual information, we envision our framework contributing to more accurate and efficient predictive models, ultimately improving patient outcomes and reducing healthcare costs.

\section*{Declaration of competing interest}

The authors declare that they have no known competing financial interests or personal relationships that could have appeared to influence the work reported in this paper.

\section*{Acknowledgements}
The authors would like to thank the CCHC cohort team, particularly Rocío Uribe who recruited and interviewed the participants. Marcela Morris, BS, and their teams for laboratory and data support respectively; Norma Pérez-Olazarán, BBA, and Christina Villarreal, BA for administrative support; Valley Baptist Medical Center, Brownsville, Texas, for providing us space for our Center for Clinical and Translational Science Clinical Research Unit is located; and the community of Brownsville and the participants who so willingly participated in this study in their city. This study was funded in part by Center for Clinical and Translational Sciences, National Institutes of Health Clinical and Translational Award grant no. UL1 TR000371 from the National Center for Advancing Translational Sciences.

\FloatBarrier

\section*{References}

 \bibliographystyle{elsarticle-num} 
 \bibliography{cas-refs}

 \appendix
\subsection{Models Pseudocodes}
\label{sec:appendix_a}
This appendix provides the pseudocode for the models used in our study. This facilitates a better understanding of their implementation and functionality. Each algorithm is described step by step, ensuring that all necessary components and operations are included.

\textbf{Algorithm 1: GVAE Model}\\
The Graph Variational Autoencoder (GVAE) is designed to encode graph features into a latent space and then decode them back to reconstruct the graph. This process is essential for learning meaningful graph representations, which are used in addition to downstream tasks.

\begin{algorithm}
\caption{GVAE Model}
\label{alg:GVAE}
\begin{algorithmic}[1]
\Require Input features $\mathbf{X}$, adjacency matrix $\mathbf{A}$

\State \textbf{Encoder:}
\State \quad $\mathbf{H} \leftarrow \text{ReLU}(\text{GCNConv}_1(\mathbf{X}, \mathbf{A}))$
\State \quad $\mathbf{H} \leftarrow \text{GCNConv}_2(\mathbf{H}, \mathbf{A})$
\State \quad $\boldsymbol{\mu} \leftarrow \text{Dense}_{\text{mean}}(\mathbf{H})$
\State \quad $\boldsymbol{\sigma} \leftarrow \text{Dense}_{\text{variance}}(\mathbf{H})$
\State \quad $\epsilon \leftarrow \mathcal{N}(0, 1)$
\State \quad $\mathbf{z} \leftarrow \boldsymbol{\mu} + \text{softplus}(\boldsymbol{\sigma}) \cdot \epsilon$

\State \textbf{KL Loss:}
\State \quad $\text{KL loss} \leftarrow -0.5 \sum (1 + \log(\boldsymbol{\sigma}^2) - \boldsymbol{\mu}^2 - \boldsymbol{\sigma}^2)$

\State \textbf{Decoder:}
\State \quad $\hat{\mathbf{A}} \leftarrow \mathbf{z} \mathbf{z}^T$
\State \quad $\hat{\mathbf{X}} \leftarrow \text{Dense}_{\text{features}}(\mathbf{z})$

\State \textbf{Return} $\mathbf{z}$, $\hat{\mathbf{A}}$, $\hat{\mathbf{X}}$, KL loss

\State \quad Adjust edges based on their importance:
\State \quad \quad Remove edges if accuracy increases, add edges if accuracy decreases
\State \textbf{Return} Adjusted adjacency matrix $\hat{\mathbf{A}}$

\end{algorithmic}
\end{algorithm}

\textbf{Algorithm 2: Training GVAE}\\
This algorithm outlines the training process for the GVAE model. It iterates over a specified number of epochs, optimizing the model based on reconstruction loss and KL divergence. The goal is to ensure that the GVAE effectively learns to represent the input graph data.

\begin{algorithm}
\caption{Training GVAE}
\label{alg:Training_GVAE}
\begin{algorithmic}[1]
\Require Input features $\mathbf{X}$, adjacency matrix $\mathbf{A}$, epochs $n$

\State \textbf{GVAE Training:}
\For{epoch $= 1$ to $n$}
    \State $\mathbf{z}, \hat{\mathbf{A}}, \hat{\mathbf{X}}, \text{KL loss} \leftarrow \text{GVAE}(\mathbf{X}, \mathbf{A})$
    \State $\text{Reconstruction loss} \leftarrow \text{loss\_fn}(\hat{\mathbf{A}}, \mathbf{A})$
    \State $\text{Total loss} \leftarrow \text{Reconstruction loss} + \text{KL loss}$
    \State Optimize GVAE with total loss
\EndFor

\end{algorithmic}
\end{algorithm}

\noindent\textbf{Algorithm 3: GNN Model}\\
The Graph Neural Network (GNN) model is tailored for processing graph-structured data. It includes several layers, such as graph convolutional layers and feedforward neural networks, to transform and analyze the input graph features.

\begin{algorithm}
\caption{GNN Model}
\label{alg:GNN}
\begin{algorithmic}[1]
\Require Graph information $(\mathbf{X}, \mathbf{A}, \mathbf{W})$, number of classes $C$, hidden units $\mathbf{h}$, dropout rate $d$

\State \textbf{Initialize:}
\State Node features $\mathbf{X}$
\State Edges $\mathbf{A}$
\State Edge weights $\mathbf{W} \gets \text{normalized weights}(\mathbf{A})$

\State $\text{Preprocess} \gets \text{FFN}(\mathbf{h}, d)$
\State $\text{GCNConv}_1 \gets \text{GCNConv}(32, \text{activation}=\text{'relu'})$
\State $\text{GCNConv}_2 \gets \text{GCNConv}(16, \text{activation}=\text{'relu'})$
\State $\text{Postprocess} \gets \text{FFN}(\mathbf{h}, d)$
\State $\text{Compute\_logits} \gets \text{Dense}(C, \text{activation}=\text{None})$

\Procedure{GNN}{$\mathbf{X}, \mathbf{A}$}
    \State $\mathbf{X} \gets \text{Preprocess}(\mathbf{X})$
    \State $\mathbf{X} \gets \text{ReLU}(\text{GCNConv}_1([\mathbf{X}, \mathbf{A}]))$
    \State $\mathbf{X} \gets \text{GCNConv}_2([\mathbf{X}, \mathbf{A}])$
    \State $\mathbf{X} \gets \text{Postprocess}(\mathbf{X})$
    \State \Return $\text{Compute\_logits}(\mathbf{X})$
\EndProcedure

\end{algorithmic}
\end{algorithm}

\textbf{Algorithm 4: Laplacian Regularized Graph Neural Network (LR-GNN)}\\
\noindent The Laplacian Regularized GNN (LR-GNN) incorporates a regularization term based on the graph Laplacian. This regularization encourages the model to produce smooth node representations, ensuring that similar nodes have similar representations.

\begin{algorithm}
\caption{Laplacian Regularized Graph Neural Network (LR-GNN)}
\label{alg:lr-gnn}
\begin{algorithmic}[1]
\Require Graph features $\mathbf{X}$, Adjacency matrix $\mathbf{A}$, Degree matrix $\mathbf{D}$, Laplacian regularization weight $\lambda$

\State \textbf{LR-GNN Training:}
\For{epoch $= 1$ to $n$}
    \State $\mathbf{y}_{\text{pred}} \gets \text{LR-GNN}(\mathbf{X})$
    \State $\mathbf{L} \gets \mathbf{D} - \mathbf{A}$
    \State $\mathbf{y}_{\text{pred\_vertices}} \gets \mathbf{L} \cdot \mathbf{y}_{\text{pred}}$
    \State $\text{laplacian\_loss} \gets \text{mean}((\mathbf{y}_{\text{pred}} - \mathbf{y}_{\text{pred\_vertices}})^2)$
    \State $\text{total\_loss} \gets \lambda \cdot \text{laplacian\_loss} + \text{lr-gnn\_losses}$
    \State $\text{apply\_gradients}(\text{total\_loss})$
    \State \Return $\mathbf{y}_{\text{pred}}$
\EndFor

\end{algorithmic}
\end{algorithm}

\section{Computational Complexity Analysis}
\label{sec:complex_analysis}

Training machine learning models, such as GVAE and GNN, requires a comprehensive understanding of the associated computational complexity and resource demands. These models are essential in various domains because of their ability to capture relevant relationships and structures within graph data, making them highly relevant for tasks that involve network analysis, molecular modeling, social network analysis, and healthcare data.

\vspace{-10pt}
\subsection{GVAE Complexity}

The model's computational complexity is $O(N^2 \cdot F')$, where $N$ is the number of nodes and $F'$ is the number of hidden units, indicating that complexity mainly depends on the square of the number of nodes and the hidden units.

\vspace{-10pt}
\subsection{GNN Complexity}

The overall complexity of the GNN model can be summarized as: $O(N \cdot F^2 + N^2)$, where $N$ is the number of nodes in $\mathbf{\mathcal{G}}$ and, $F$ is the number of features per node. This indicates that computational complexity depends on the number of nodes, input features, output features, and edges in the graph.

Training these models involves substantial computation resources, especially when considering a large number of epochs. For instance, the GVAE model is trained for 3000 epochs and the GNN model is trained for 12000 epochs. Given the size and complexity of the models, the training involves extensive computational load and time.

\end{document}